\documentclass[conference]{IEEEtran}
\usepackage{amsthm,amsmath,amssymb,amsfonts}
\usepackage{mathrsfs}

\usepackage{algorithmic}
\usepackage{graphicx}
\usepackage{textcomp}
\usepackage[misc]{ifsym}

\usepackage{multirow}
\usepackage{booktabs}
\usepackage{bbding}

\usepackage{color}
\usepackage{xcolor}

\usepackage{etoolbox}
\makeatletter
\@ifundefined{color@begingroup}%
  {\let\color@begingroup\relax
   \let\color@endgroup\relax}{}%
\def\fix@ieeecolor@hbox#1{%
  \hbox{\color@begingroup#1\color@endgroup}}
\patchcmd\@makecaption{\hbox}{\fix@ieeecolor@hbox}{}{\FAILED}
\patchcmd\@makecaption{\hbox}{\fix@ieeecolor@hbox}{}{\FAILED}
\usepackage{biblatex}
\usepackage[]{hyperref}
\newcommand{\TheName}{\textbf{CUPre}}

\newcommand{\wqz}[1]{\textcolor{black}{#1}}

\def\BibTeX{{\rm B\kern-.05em{\sc i\kern-.025em b}\kern-.08em
    T\kern-.1667em\lower.7ex\hbox{E}\kern-.125emX}}
\begin{document}
\title{\TheName{}: Cross-domain Unsupervised Pre-training for Few-Shot Cell Segmentation}
\author{Weibin~Liao$^{1}$,~Xuhong~Li$^1$~Qingzhong~Wang$^1$,~Yanwu~Xu$^2$,~Zhaozheng~Yin$^3$,~and Haoyi~Xiong\Envelope$^1$
\\
$^1$ Big Data Lab, Baidu, Inc., Beijing, China\\
$^2$ HDMI Lab, South China University of Technology, Guangdong, China\\
$^3$ Department of Computer Science, Stony Brook University, Stony Brook, NY}

\maketitle

\begin{abstract}
%Cell segmentation with 
%The Deep neural network (DNN) has become an indispensable tool to analyze the structures and behaviors of cells in pathological microscopic images. 
While pre-training on object detection tasks, such as Common Objects in Contexts (COCO)~\cite{lin2014microsoft}, could significantly boost the performance of cell segmentation, it %requires 
still consumes on massive fine-annotated %\wqz{pathological microscopic} 
cell images~\cite{edlund2021livecell} with bounding boxes, masks, and cell types for every cell in every image, to fine-tune the pre-trained model.~%, subject to the instance segmentation task in the visual domain of cells. 
%In this paper, we investigate the task of few-shot cell segmentation to lower the cost of annotation.
%
To lower the cost of annotation, this work considers the problem of pre-training DNN models for few-shot cell segmentation, where massive unlabeled cell images are available but only a small proportion is annotated. Hereby, we propose \emph{\underline{C}ross-domain \underline{U}nsupervised \underline{Pre}-training}, namely \TheName{}%~\footnote{In Latin, the word ``cupre'' means ``copper'' in English.}
, transferring the capability of object detection and instance segmentation for common visual objects (learned from COCO) to the visual domain of cells using unlabeled images. 
Given a standard \emph{COCO} pre-trained network with backbone, neck, and head modules, \TheName{} adopts an \emph{alternate multi-task pre-training} (AMT$^2$) procedure with two sub-tasks --- in every iteration of pre-training, AMT$^2$ first trains the backbone with cell images from multiple cell datasets via unsupervised \emph{momentum contrastive learning (MoCo)~\cite{he2020momentum}}, and then trains the whole model with vanilla \emph{COCO} datasets via instance segmentation. %\wqz{Obviously, AMT$^2$ pre-training enables the model to have the capability of segmentation and the knowledge of microscopic images.}
%Given the model pre-trained by the AMT$^2$ procedure, 
%After that, we fine-tune the entire model on various cell instance segmentation tasks using a few annotated images.
After pre-training, \TheName{} fine-tunes the whole model on the cell segmentation task using a few annotated images. We carry out extensive experiments to evaluate \TheName{} using LIVECell \cite{edlund2021livecell} and BBBC038 \cite{caicedo2019nucleus} datasets in few-shot instance segmentation settings. The experiment shows that \TheName{} can outperform existing pre-training methods, achieving the highest average precision (AP) for few-shot cell segmentation and detection. Specifically, when using only 5\% annotated images from LIVECell for training, \TheName{} achieves 41.5\% $\mathbf{AP_{bbox}}$ versus 40.0\% and 8.3\% obtained by the \emph{COCO} pre-trained and MoCo-based LIVECell pre-trained models respectively.
%
%\emph{COCO} pre-trained  model only obtains 40.0\% $\mathbf{AP_{bbox}}$ and MoCo pre-trained model only achieves 8.3\% $\mathbf{AP_{bbox}}$. 
External validation using out-of-distribution datasets further confirms the generalization superiority of \TheName{}, which outperforms baselines with 10\% higher $\mathbf{AP_{bbox}}$ when using 5\% BBBC038 images for few-shot learning.
%BBBC038 dataset further confirms the superiority of \TheName{} in out-of-distribution generalization, 
%for example, 53.8\% vs. 43.7\% $\mathbf{AP_{bbox}}$ using 5\% annotated images.
\end{abstract}


\section{Introduction}
\label{sec:introduction}

\begin{figure}[htb]
\centering
\includegraphics[width=0.48\textwidth]{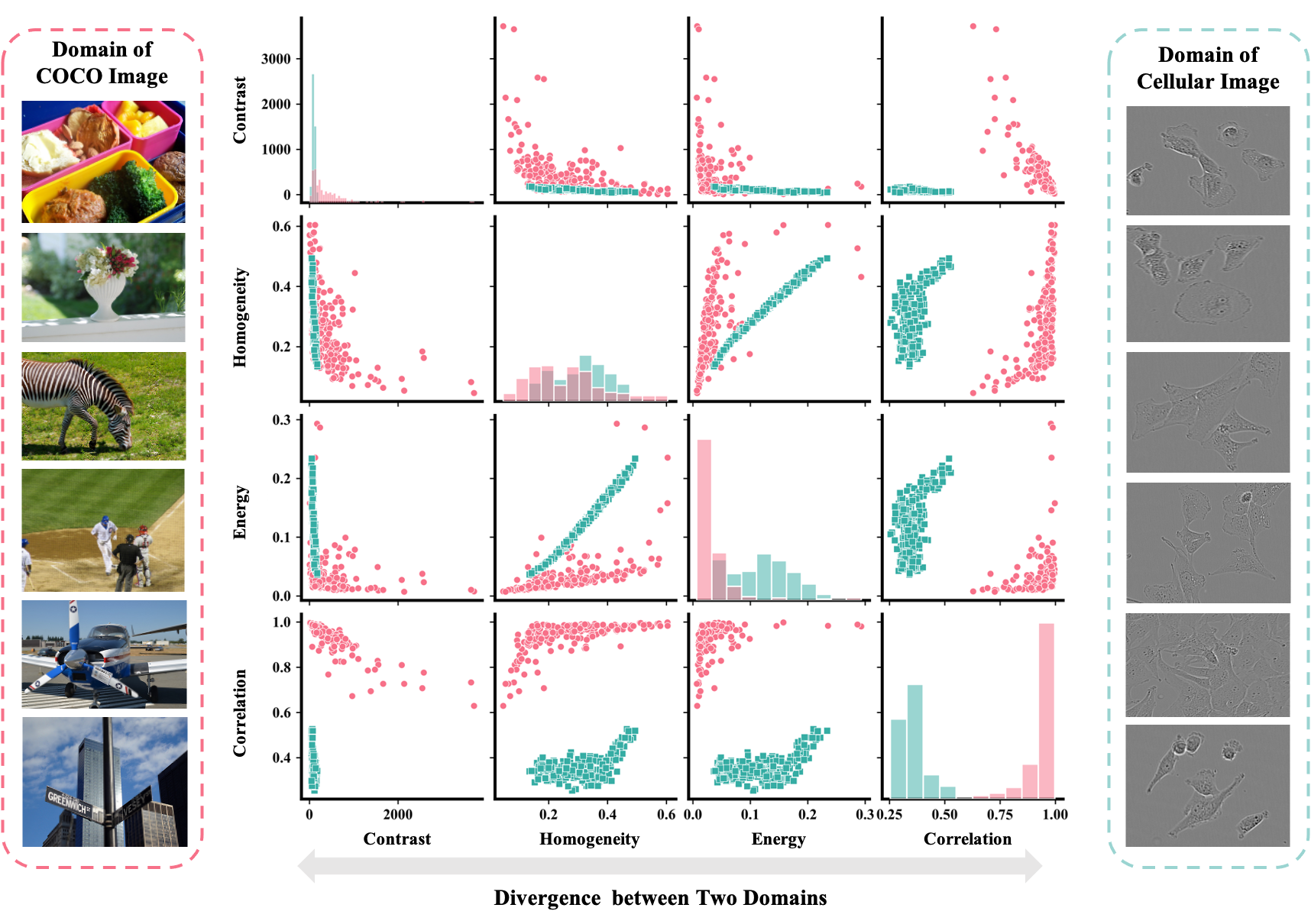} %\vspace{-5mm}
\caption{The divergence between visual domains of \emph{COCO} \cite{lin2014microsoft} and LiveCell~\cite{edlund2021livecell}. Features of contrast, homogeneity, energy and correlation are extracted from 200 images randomly selected from both datasets.}
\label{fig:gap between two gomains}
\vspace{-5mm}
\end{figure}

%\IEEEPARstart{D}{eep} neural network (DNN) for computer vision is increasingly being 
%used for a variety of tasks in biological image analysis, including nuclear and cell segmentation \cite{van2016deep,hollandi2020nucleaizer}. 
%%It 
%\wqz{Deep models have} achieved high and robust accuracy for cell instance segmentation, %task of cell with a robust identification and segmentation solution, 
%serving as the cornerstone of cell analysis to allow quantification of the effects of interventions on cell count, proliferation, morphology, migration, and cell interactions \cite{edlund2021livecell}.

%However, for a DNN to produce good results, it %first 
%requires training with high-quality datasets representing the breadth of the problem to be solved, leading to extensive, tedious and expensive manual annotation, especially in biomedical imaging, which needs to be performed by experienced domain experts. 

%To address this challenge, 
% a lot of recent works have been devoted to \emph{few-shot learning} study for lower the cost for manual annotation \cite{NIE2020539}.
%\emph{few-shot learning} has become an area of growing interest for lower the cost of manual annotation \cite{NIE2020539}.

%\IEEEPARstart{C}{ellular} 
Cellular assays are an accessible medium to obtain physiologically relevant data from images, which allow the quantification of intervention effects on cell counts, proliferation, morphology, migration, cell interactions~\cite{edlund2021livecell} and the translational research~\cite{greenwald2022whole}. While these assays ultimately rely on robust identification and segmentation, especially, at the granularity of individual cells, the rise in popularity of deep neural networks (DNNs) offers a potential solution to this problem~\cite{van2016deep,kermany2018identifying}. %and indeed, DNNs can learn and adapt to identify and segment objects of enormous variety. 
However, a DNN usually needs to consume a huge mount of fine-annotated cell images~\cite{van2016deep} to deliver good results. To the end, few-shot learning~\cite{cui2020unified,ouyang2022self} becomes desired to train cell segmentation models with a few annotated images.

%As massive unlabeled cell images have been made publicly available~\cite{edlund2021livecell}, the \emph{pre-training} and \emph{fine-tuning} pipeline~\cite{hu2022pushing} becomes a practical option to leverage both labeled and unlabeled images for few-shot segmentation. 

% To reduce the cost of manual annotation of dataset, the pre-training efforts for CNN are widely appearing in various fields. The primary goal of pre-training CNN is to learn transferable prior knowledge from mostly unlabeled data, which can be generalized to downstream tasks with a quick fine-tuning step. \cite{navarin2018pre, hu2019pre, hu2019strategies}

On the other hand, some recently published works achieve remarkable performance on cell segmentation \cite{edlund2021livecell,dietler2020convolutional,stringer2021cellpose}.
Among these efforts, Edlund et al. \cite{edlund2021livecell} present the LIVECell, a new dataset composed of label-free, phase-contrast images of 2D cell culture with substantial amounts of annotations, and they utilize DNNs to perform cell instance segmentation for label-free single-cell study. 
To achieve %the best 
better performance, authors adopt a DNN pre-trained using the Common Objects in Contexts (COCO) dataset \cite{lin2014microsoft} and fine-tune the model with annotated cell images. 
Benefiting from the huge amount of annotated images (2,500,000 labeled instances in 328,000 images) for fine-tuning, 
such COCO-based pre-training has %been proven to be 
shown effectiveness~\cite{edlund2021livecell}. %including medical images. 
However we believe the performance of pre-training could be further improved, as
%it is limited to handle few-shot cell segmentation due to two reasons
%
the domain divergence between natural images of \emph{COCO} \cite{lin2014microsoft} and microscopic images \cite{raghu2019transfusion} in  LIVECell \cite{edlund2021livecell} remains large (as shown in Fig~\ref{fig:gap between two gomains}).
To relieve the domain divergence and boost the performance on few-shot cell segmentation, %some biomedical image analysis 
one possible method is to leverage self-supervised learning (SSL) to pre-train backbones of DNN using  unlabeled images~\cite{felfeliyan2022self,liao2022muscle}.~% have moved towardswhich trains a DNN using a large number of . 
%to boost the performance of DNN models through learning visual representations from a large number of cell images without the use of labels.
%More precisely, 
Compared to the COCO-based pre-training, SSL allows a backbone network %of DNN model first 
to learn good visual representations from unlabeled images %usually with pretext tasks, 
and is capable of transferring the knowledge to downstream tasks in the same domain. However, SSL doesn't pre-train the neck or head of the DNN -- two critical components for segmentation networks, while the COCO-based solution preserves the capacity of instance segmentation learned from common visual objects. 
%DNN with a small number of well-annotated images.
% While SSL is proven to assist in providing knowledge of cell to the backbone network, for other components of DNN, such as the task-specific neck and head, it is powerless.
%\wqz{Though SSL enables the backbone network to obtain domain knowledge, the entire model lacks task-related capability. To address this issue, task-specific pre-training on \emph{COCO} is promising, referring to the proposed alternate multi-task pre-training (AMT$^2$) procedure.}
%
%In summary, collecting large amounts of fine-annotated cell images is expensive and time-consuming. Therefore, it is essential to build cross-domain pre-training models that possess transferability to quickly adapt to the domain of cell images by using only a few annotated samples.
In this way, there is a need to transfer the capability of object detection and instance segmentation for common visual objects (learned from COCO) to the visual domain of cells using unlabeled images.
%
% In summary, \emph{COCO} \cite{lin2014microsoft} pre-training algorithm generates task-relevant representations but does not provide knowledge related to visual domain of cells, and SSL solution could generate image representations of cells but do not provide task-relevant knowledge.
% Therefore, it is crucial to build a new pre-training architecture crossing image and task domains to pre-train model thus lower cost of manual annotation. 
% Specifically, it learns representations of image form unlabeled visual domain of cell and learns representations of task from other visual domain with task-specific annotation, to generate a model that captures the knowledge of the target image/task domain and performs well on few-shot downstream task.
To achieve the goal, we proposed \TheName{}, a \emph{\underline{c}ross-domain \underline{u}nsupervised \underline{pre}-training} for few-shot cell segmentation, where
\wqz{\TheName{} equips Mask R-CNN models~\cite{he2017mask} with both instance segmentation capacity and visual domain knowledge of cells through \emph{alternate multi-task pre-training} (AMT$^2$).}
%Specifically, it allows DNN learn knowledge of instance segmentation for common visual objects from \emph{COCO} dataset \cite{lin2014microsoft}, and then transfers the capability to visual domain of cells using unlabeled cell images. In this work, \TheName{} pre-trained DNN on a collection of \textbf{nine cell image datasets and \emph{COCO} dataset \cite{lin2014microsoft}}, and validated performance on \textbf{two cell instance segmentation tasks}.
% transfers the capability of object detection and instance segmentation for common visual objects (learned from COCO) to the visual domain of cells using unlabeled cell images according to two subtasks, including pre-training the backbone of DNN on a collection of \textbf{nine cell image datasets} using SSL paradigm and pre-training the whole DNN with \textbf{task-specific neck/head} on \emph{COCO} dataset using supervised solution. Then \TheName{} evaluated performance for instance segmentation task on two real-world cell image datasets. We made contributions as follows:
The contributions made in our work could be categorized in three-fold.
\begin{enumerate}
    \item In this work, we study the feasibility of pre-training DNNs for few-shot cell segmentation, where we assume massive unlabeled cell images are available with an extremely small proportion of images annotated. To the best of our knowledge, this study is the first work on the problem of cell segmentation by addressing few-shot learning, pre-training for instance segmentation networks, and domain divergence issues. 
    %To achieve the best performance, we overcome the challenge of \emph{domain divergence} between natural images and cell images, and avoid \emph{knowledge forgetting} using regularization constraints during alternate learning step.
    
    \item We present \TheName{} that uses COCO~\cite{lin2014microsoft} and multiple cell image datasets to pre-train the whole DNN in an end-to-end manner. Given a standard \emph{COCO} pre-trained network composed of a ResNeSt-200 backbone~\cite{zhang2020resnest}, a neck and a head based on Cascade Mask R-CNN~\cite{cai2019cascade}, \TheName{} adopts an AMT$^2$ procedure with two sub-tasks --- in every iterate of pre-training, AMT$^2$ first trains the backbone with cell images from multiple datasets via unsupervised \emph{momentum contrastive loss} and then trains the while model (backbone, neck and head modules) with vanilla \emph{COCO} datasets via instance segmentation. Finally, \TheName{} fine-tunes the model on the cell segmentation task using a few annotated images.

    %Given a DNN for instance segmentation task, \TheName{} first warms up the whole model, including backbone, neck and head using instance segmentation task of COCO, and then it performs AMT$^{2}$ process for knowledge transfer with two sub-tasks in a loop, i.e., AMT$^2$ trains the backbone with multiple cell datasets using unsupervised momentum contrastive learning and then trains the task-specific neck and head with \emph{COCO} \cite{lin2014microsoft} via instance segmentation task again. Finally, \TheName{} fine-tunes the entire model on down-stream instance segmentation tasks using few-shot settings with a small number of annotated samples.
    
    % to transfer the capability learned from \emph{COCO} domain to cell. To better exploit the value, we overcome the challenge of domain divergence between natural images and biomedical images using an AMT$^{2}$ procedure, and avoid knowledge forgetting using regularization constraints during alternate learning step.
    % \item We explore a major component of this work is deficiency analysis of \emph{COCO} \cite{lin2014microsoft} and SSL pre-training algorithms, and design a novel pipelines to tackle this challenge. Experimental results show that \TheName{} learns representations related to instance segmentation tasks of cells, and based on them, \TheName{} achieves the best performance.
    \item We carry out extensive experiments to evaluate \TheName{} using LIVECell \cite{edlund2021livecell} and BBBC038 \cite{caicedo2019nucleus} datasets in few-shot instance segmentation settings. The experiment shows that \TheName{} can outperform existing pre-training methods, achieving the highest average precision (AP) for both instance segmentation and bounding box detection. For example, \TheName{} achieves 41.5\% $AP_{bbox}$ on the LIVECell-wide dataset using only 5\% annotated images, while using \emph{COCO} pre-training obtains 40.0\% $AP_{bbox}$, and self-supervised learning only achieves 8.3\% $AP_{bbox}$. External validation using the BBBC038 dataset further confirms the superiority of \TheName{} in out-of-distribution generalization, for example, 53.8\% vs. 43.7\% $AP_{bbox}$ using 5\% annotated images.

    %We dedicated our research to few-shot learning to lower the cost of manual annotation during training of DNNs for cell segmentation. Experimental results show that \TheName{} achieves remarkably better performance than other pre-training algorithms, particularly for few-shot learning. Moreover, an external validation further confirms the superiority of \TheName{}. %in out-of-distribution generalization.
\end{enumerate}

\section{Related Work}
%\subsection{Deep Learning for Medical Image Analysis}
%Deep learning \cite{kermany2018identifying} has made tremendous success in cell image analysis, such as cell counting, classification, segmentation, detection, and morphometry. 
%In this section, we present some related works on cell image analysis~\cite{kermany2018identifying}.
%Though deep learning solutions significantly boost the performance of cell analysis, it relies on a large number of manually annotated datasets due to the supervised learning paradigm. Here we summarize the existing works from the problem and methodological perspectives.

\paragraph{Cell Instance Segmentation}
Automatic cell detection and counting in microscopy images have been  studied for years \cite{faustino2009automatic,lu2015improved,wahlby2004combining}. 
U-Net \cite{ronneberger2015u} is adopted to analyze neuronal structures in electron microscopy images, but U-Net is limited and cannot distinguish each cell separately in one image.
To segment cells accurately, Oda et al. \cite{oda2018besnet} further estimate the boundaries of cells in a segmentation map.
Yi et al. \cite{yi2019attentive} propose an attentive instance segmentation model which combines object detection and segmentation architectures. 
Payer et al. \cite{payer2019segmenting} propose a loss function that introduces instance information into a semantic segmentation model to boost the performance of instance segmentation.
However, these methods are trained in a supervised learning manner,  relying on large-scale pixel-level annotations. For images with dense cells, the annotations are time-consuming and expensive. 
Moreover, domain shift occurs when using different types of microscopes; hence, there are many conditions in cell instance segmentation \cite{nishimura2021weakly}, and it is impossible to consider each condition. In this case, few-shot cell instance segmentation is promising.
%In particular, there are too many combinations of cell type, microscope type, and culture condition to be able to annotate cell regions for each condition at the pixel level \cite{nishimura2021weakly}.
%In summary, cell instance segmentation with a few-shot learning paradigm is crucial to aid in pathological analysis.

\paragraph{Self-supervised Learning for Cell Images}
%Self-supervised learning is an especially promising approach, allows networks to leverage unlabeled training data and learn to extract meaningful representations without any type of manual annotation or data curation.
%to enable models to be trained on unlabeled images in pathology. 
Pre-train-fine-tune paradigm dominates the transfer learning field of computer vision, and many researchers have adopted it to analyze cell images.
%Related studies also act on cell images. 
Lu et al. \cite{lu2019learning} employ an unsupervised image inpainting method to learn better representations for single-cell microscopy image analysis. %improve the downstream task of identifying protein sub-cell localization classes at the single-cell level with representation learned by SSL.
%Based on the pretext task of SSL, 
Zheng et al. \cite{zheng2018fast} apply K-means clustering to separate the background and foreground of blood smear images and then segment white blood cells (WBCs) using shape-based concavity analysis.
%In work of 
Meanwhile, Yamamoto et al. \cite{yamamoto2019automated} design a pretext task that employs nucleus structures of cells in both high and low magnification images %as well as the structural pattern of cells analyzed in low magnification images,
to learn interpretable representations.
%and indicated that representations learned by SSL were also unique in that they were able to be understood by pathologists. 
Over the past three years, general SSL techniques are continually developed, and more SSL approaches have been applied to analyze pathology images for cells \cite{gildenblat2019self,tellez2019neural}.

\paragraph{Comparisons with Relevant Works}
The most relevant works are~\cite{dawoud2020few,xun2021scellseg}. %from the problem and solution perspectives, respectively. 
Compared to~\cite{dawoud2020few}, we both focus on few-shot learning for cell segmentation to lower the cost of annotation. 
In~\cite{dawoud2020few}, meta-learning with three loss functions for cell segmentation is proposed, and the model focuses on semantic segmentation, which cannot provide a bounding box and category for each cell. 
In contrast, \TheName{} adopts pre-training to obtain prior domain and task knowledge, benefiting downstream tasks of cell instance segmentation.
Though Xun et al. \cite{xun2021scellseg} also use pre-training to improve performance, they only use contrastive learning to obtain domain knowledge. Whereas, the proposed \TheName{} employs not only contrastive learning to initialize the backbone network but also supervised learning on instance segmentation tasks using \emph{COCO} dataset to initialize neck and head networks, allowing the entire model to have both domain and task knowledge. Thus, it can be easily transferred to downstream tasks with only a few annotated images for fine-tuning.
%Compared to~\cite{xun2021scellseg}, both of us intended to boost performance of DNN using pre-training solution. However, ~\cite{xun2021scellseg} adopted contrastive learning algorithom to pre-train model and proposed to also employ unlabeled data in fine-tuning strategies, then established a fine-tuning pipeline for improved extraction and utilization of style features.  \TheName{} combines natural images and cell images to achieve cross-domain, edge-based pre-training solution, and it also makes effort on various fine-tuning strategies described in Sec~\ref{sssec:ablation study}.

\section{\TheName{}: Methodologies and Configurations}~\label{sec:method}
\begin{figure*}
\centering
\includegraphics[width=0.9\textwidth]{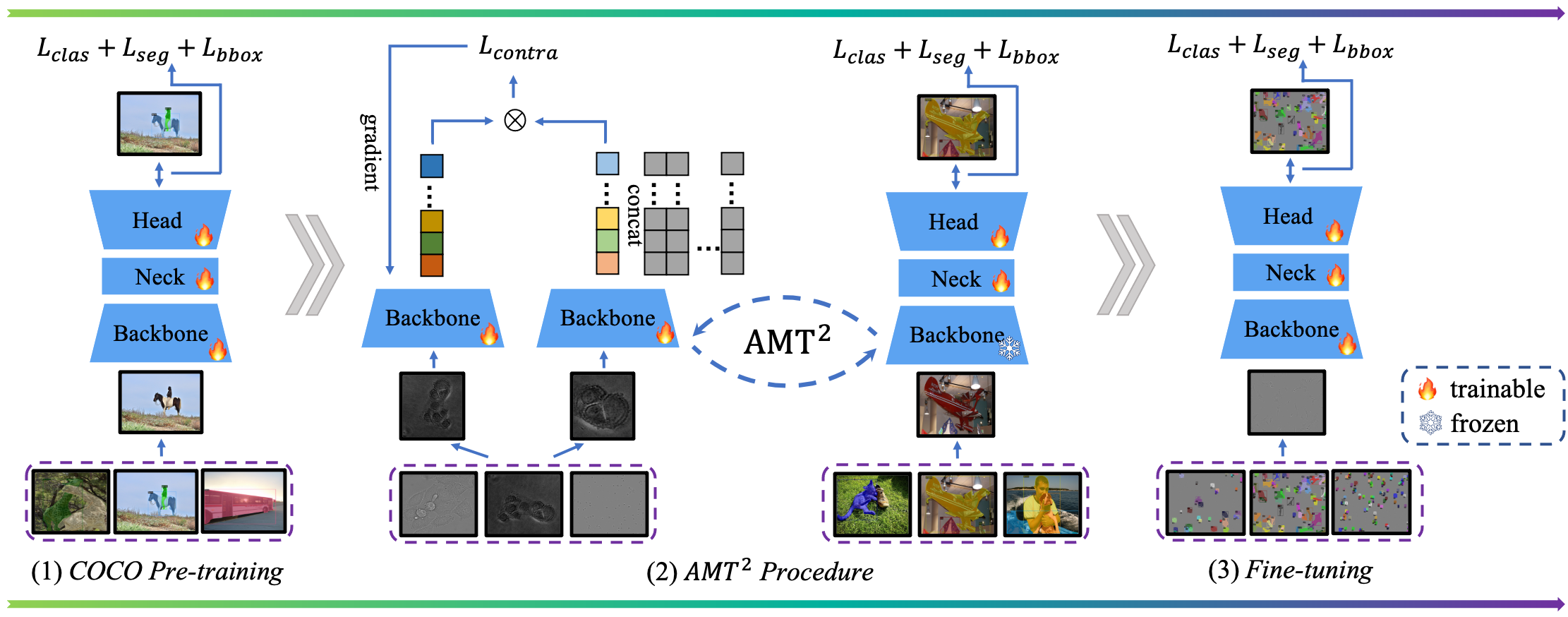}
\caption{\TheName{} consisting of three steps: (1) \emph{COCO Pre-training}, (2) \emph{Alternating Multi-Task Training} (AMT$^2$) with two subtasks--instance segmentation on \emph{COCO} datasets and self-supervised training~\cite{he2020momentum} on unlabeled cell images, and (3) \emph{Fine-tuning} on a few annotated cell images.}
\label{fig:framework}
\vspace{-5mm}
\end{figure*}
For few-shot cell segmentation, we propose the \emph{\underline{C}ross-domain \underline{U}nsupervised \underline{Pre}-training} (\TheName{}) pipeline to transfer the capability of instance segmentation from common visual objects to the domain of cell images. In this section, we first present the datasets and configurations for pre-training, then introduce the main pipeline of \TheName{}. We also present the design of key algorithms in details.

\subsection{Datasets Preparation for \TheName{} Pre-training}
To transfer the instance segmentation capability of natural images to cell images, \TheName{} collects and aggregates \emph{eleven open-source datasets} reported in Table.~\ref{tab:dataset}, including images of both common visual objects and cells, for pre-training. Specifically, \TheName{} leverages the full dataset of COCO, including both images and the labels for instance segmentation. Furthermore, to set up the pre-training for few-shot cell segmentation, \TheName{} uses only the images from the training sets of cell image datasets, including LIVECell-Unlabeled (created from LIVECell \cite{edlund2021livecell}), EVICAN \cite{3.AJBV1S_2019}, BBBC038 \cite{caicedo2019nucleus}, and the datasets published by the Cell Tracking Challenge \cite{ulman2017objective} consisting of DIC-C2DH-HeLa, Fluo-C2DL-MSC, Fluo-N2DH-GOWT1, Fluo-N2DH-SIM+, Fluo-N2DL-HeLa, PhC-C2DH-U373 and PhC-C2DL-PSC. In this way, a total of 11 datasets covering both natural images and cell images are used for \TheName{} Pre-training.

%While these datasets have offered a total of 118,287 nature images and 12,470 cell images covering a large natural image dataset with manual annotation of instance segmentation, \emph{COCO} \cite{lin2014microsoft} and ten cell images including LIVECell \cite{edlund2021livecell}, EVICAN \cite{3.AJBV1S_2019}, BBBC038 \cite{caicedo2019nucleus}, and the datasets published by the Cell Tracking Challenge \cite{ulman2017objective} consisting of DIC-C2DH-HeLa, Fluo-C2DL-MSC, Fluo-N2DH-GOWT1, Fluo-N2DH-SIM+, Fluo-N2DL-HeLa, PhC-C2DH-U373 and PhC-C2DL-PSC. 

\subsection{\TheName{}: Overall Pipeline}
As shown in Fig.~\ref{fig:framework}, \TheName{} adopts a three-step pipeline to pre-train DNN models for few-shot cell segmentation. Specifically, \TheName{} leverages a Cascade Mask R-CNN \cite{cai2019cascade} model with the backbone of ResNeSt-200 \cite{zhang2020resnest} for object detection and instance segmentation in images.
In the pipeline, \TheName{} (1) pre-trains the model on \emph{COCO} \cite{lin2014microsoft} datasets to warm up the model with the capacity of instance segmentation for common visual objects. Then, given images from \emph{COCO} and cell image datasets, \TheName{} (2) proposes the \emph{AMT$^2$ Procedure} to train each module of the DNN, including the backbone, neck, and head modules, to adapt the visual domain of cell images while preserving the instance segmentation capacity of common visual objects. Finally, \TheName{} (3) fine-tunes the whole DNN subject to the cell segmentation task using a few annotated cell images.

%Specifically, \TheName{} allows the backbone of DNN to learn cell-relevant knowledge using the self-supervised contrastive learning paradigm with a large number of unlabeled cell images, and then tunes the neck and head of DNN to adapt the whole model using instance segmentation paradigm with the labeled \emph{COCO} \cite{lin2014microsoft} dataset.

\begin{table}[]
\caption{An Overview of Eleven Publicly Available Cell Image Datasets. * We remove labels for images in the training dataset of LIVECell to synthesize LIVECell-Unlabeled.}
\label{tab:dataset}
%\resizebox{0.5\textwidth}{!}{
%\begin{tabular}{llcr}
\centering
\begin{tabular}{llr}
\toprule
 & \multicolumn{1}{c}{\textbf{Datasets}} %& \textbf{Train/Valid/Test} 
 & \multicolumn{1}{c}{\textbf{\# Images}} \\
\midrule
\multirow{10}{*}{Pre-training} & COCO \cite{lin2014microsoft} %& N/A 
& 118,287 \\
\cmidrule(r){2-3}
 & LIVECell-Unlabeled$^*$ \cite{edlund2021livecell} %& N/A 
 & 3,253 \\
 & EVICAN \cite{3.AJBV1S_2019} %& N/A 
 & 4,738 \\
 & DIC-C2DH-HeLa \cite{ulman2017objective} %& N/A 
 & 166 \\
 & Fluo-C2DL-MSC \cite{ulman2017objective} %& N/A 
 & 96 \\
 & Fluo-N2DH-GOWT1 \cite{ulman2017objective} %& N/A 
 & 184 \\
 & Fluo-N2DH-SIM+ \cite{ulman2017objective} %& N/A 
 & 215 \\
 & Fluo-N2DL-HeLa \cite{ulman2017objective} %& N/A 
 & 184 \\
 & PhC-C2DH-U373 \cite{ulman2017objective} %& N/A 
 & 230 \\
 & PhC-C2DL-PSC \cite{ulman2017objective} %& N/A 
 & 600 \\
\midrule
\multirow{6}{*}{Fine-tuning} & \multirow{3}{*}{LIVECell  \cite{edlund2021livecell}} %& Train 
& 3,253 (Train) \\
 &  %& Valid 
& 570 (Valid) \\
 &  %& Test 
& 1,564 (Test) \\
\cmidrule(r){2-3}
 & \multirow{3}{*}{BBBC038 \cite{caicedo2019nucleus}} %& Train 
 & 536 (Train) \\
 &  %& Valid 
 & 67 (Valid) \\
 &  %& Test 
 & 67 (Test)\\
\bottomrule
\end{tabular}
%}
\vspace{-5mm}
\end{table}

\subsection{COCO Pre-training}\label{sssec:coco pretrain}
To warm up the DNN with the capacity of instance segmentation (for common visual objects), \TheName{} follows the settings of Cascade Mask-RCNN \cite{cai2019cascade} on the \emph{COCO} \cite{lin2014microsoft} dataset. Specifically, this step consists of two parts as follows.

\subsubsection{Model Initialization} To obtain better generalization performance, \TheName{} first builds up a Cascade Mask R-CNN model with a backbone of ResNeSt-200, based on the implementation of Detectron2-ResNeSt~\cite{zhang2020resnest}. Then, \TheName{} initializes the model with random weights using Kaiming's strategy~\cite{he2015delving}. This step provides \TheName{} a robust set of initial weights, regarding the nonlinear properties of rectifiers, and enables \TheName{} to train deep models from scratch while working well on a wide range of deep architectures, such as the Cascade Mask R-CNN model on top of ResNeSt-200. 

\subsubsection{Training with Instance Segmentation Tasks} \TheName{} employs a standard instance segmentation training procedure to train the Cascade Mask R-CNN model on \emph{COCO} datasets. Given an image $I$, the model outputs the object class $clas$, bounding boxes $box$ and masks $seg$ as follows.
\begin{equation}
    clas, box, seg = \operatorname{\mathit{Head}}(
    \operatorname{\mathit{Neck}}(
    \operatorname{\mathit{Back}}(I)))\label{eq:model}
\end{equation}
where $\operatorname{\mathit{Back}}(\cdot)$, $\operatorname{\mathit{Neck}}(\cdot)$ and $\operatorname{\mathit{Head}}(\cdot)$ refer to the backbone, neck, and head modules of the Cascade Mask R-CNN model for pre-training, respectively. Specifically, as shown in Eq.~\ref{eq:model}, the model first leverages $\operatorname{\mathit{Back}}(\cdot)$ to extract multi-scale feature maps to adapt various objects with different scales from the \emph{COCO} image $I$. 
Then the model leverages $\operatorname{\mathit{Neck}}(\cdot)$ to improve the extracted multi-scale feature maps through fusing the high- and low-level features via a Feature Pyramid Network (FPN) \cite{lin2017feature}.
Finally the model predicts the binary masks (segmentation),  bounding boxes (detection), and object class for every object in the image $I$ using $\operatorname{\mathit{Head}}(\cdot)$.

\TheName{} trains the model in Eq~\ref{eq:model} with 4 epochs on \emph{COCO} using the standard instance segmentation loss~\cite{he2017mask}, denoted as $\mathcal{L}_{inst}$, which is composed of three components:
    \begin{equation}\label{equ:instance_loss}
        \mathcal{L}_{inst} = \mathcal{L}_{clas} + \mathcal{L}_{box} + \mathcal{L}_{seg}
    \end{equation}
where $\mathcal{L}_{clas},\mathcal{L}_{box},\mathcal{L}_{seg}$ represent the classification, box regression, and segmentation losses, respectively. %The outcome of training 

\subsection{AMT$^2$ Procedure}\label{sssection:amt procedure}
As shown in Fig.~\ref{fig:framework}, given the Cascade Mask R-CNN model warmed-up by \emph{COCO Pre-training}, \TheName{} incorporates \emph{AMT$^2$ Procedure} to transfer the instance segmentation capacity of common objects to the visual domain of cell images. The algorithm is designed as follows.

%for a DNN with the capability of object detection and instance segmentation for common visual objects (learned from \emph{COCO pre-training step}), \TheName{} transfers the capability to the visual domain of cell images by leveraging DNN to alternate learning two sub-tasks including (1) training the $\operatorname{\mathit{Back}}$ with cell images from multiple cellular datasets via unsupervised momentum contrastive loss and (2) training all $\operatorname{\mathit{Back}}$, $\operatorname{\mathit{Neck}}$ and $\operatorname{\mathit{Head}}$ with vanilla \emph{COCO} datasets via instance segmentation.

\subsubsection{AMT$^2$ Setup} Given the collection of cell images and the \emph{COCO} \cite{lin2014microsoft} dataset for pre-training in Table~\ref{tab:dataset}, AMT$^2$ splits the \emph{COCO} \cite{lin2014microsoft} dataset into 10 equal-sized subsets, replicates 10 copies of the full cell images collection, and pairs every replicate with a subset of the \emph{COCO} dataset. In this way, AMT$^2$ forms training data pairs for 10 iterations of the pre-training procedure. Specifically, in every iteration, AMT$^2$ first trains the backbone module using cell images via \emph{momentum contrastive} (MoCo) learning~\cite{he2020momentum}, and then adapts the neck and head modules using the \emph{COCO} dataset via instance segmentation tasks, respectively.

%ensure that each subset has similar number of images as the whole cell dataset used for pre-training. For MoCo \cite{he2020momentum} sub-task, \TheName{} learns the whole cell dataset in each iteration, and for instance segmentation sub-task, \TheName{} learns one non-repetitive subsets of \emph{COCO} \cite{lin2014microsoft} dataset in each iteration. Furthermore, to avoid overfitting to any sub-task, \TheName{} trains the DNN with one epoch on each iteration of learning sub-tasks.

\subsubsection{Cell Images MoCo Step} Given the collection of cell images and the Cascade Mask R-CNN model for pre-training, this step trains the backbone module of the model using the cell images via \emph{momentum contrastive} (MoCo) learning. Specifically, for every cell image $I_{c}$, the algorithm would adopt \emph{random data augmentation}, denoted as $\operatorname{\mathit{randAug}(\cdot)}$, to generates two views from the original image, i.e., 
\begin{equation}
 I_1, I_2 = \operatorname{\mathit{randAug}}({I_{c}})\ .   
\end{equation}
 Then, the algorithm randomly picks two views generated from the same image, or different images to form the training set of MoCo. When the two views are from the same original image, the algorithm labels the pair of two views as a positive pair; otherwise, the algorithm labels the pair as a negative pair. 
Latter, the algorithm extracts the representation vector of every view using the backbone module $\operatorname{\mathit{Back}}$, i.e., 
\begin{equation}
    r_{I_1} = \operatorname{\mathit{Back}}(I_1),\ r_{I_2} = \operatorname{\mathit{Back}}(I_2)\ \text{and}\ r_{I_1},r_{I_2}\in \mathbb{R}^{D_r}\ ,
\end{equation} 
where $D_r$ is the dimension of learned representations. Finally, the algorithm adopts a \emph{projection network}, denoted as $\operatorname{\mathit{Proj}(\cdot)}$, to map the learned representations extracted from the backbone modules into a low-dimensional space, i.e., 
\begin{equation}
    v_{I_{1}} = \operatorname{\mathit{Proj}}(r_{I_1}),\ v_{I_{2}} = \operatorname{\mathit{Proj}}(r_{I_2}),\ \text{and}\ v_{I_{1}}, v_{I_{2}}\in \mathbb{R}^{D_{v}}\ ,
\end{equation}
where $D_v \ll D_r $. The objective of MoCo is to align the learned representation vectors of two views from the same image and discriminate ones from the different images, such that the \emph{contrastive loss} $\mathcal{L}_{contra}$ is defined as follows.
\begin{equation}\label{equ:contra_loss}
    \mathcal{L}_{contra} %= \sum_{i \in I} \mathcal{L}_{i}
    = - %\sum_{i \in I} 
    \log \frac
    {\exp \left(v_{I_{1}} \cdot v_{I_{2}} / \tau\right)}
    {\sum_{I^{-} \in \Omega^{-}} \exp \left(
    v_{I_{1}} \cdot v_{I^{-}} / \tau
    \right)}\ ,
\end{equation}
where the $\tau\in (0,1]$ refers to a tuning parameter, and the $I^{-}\in \Omega^-$ denotes the negative samples of $I$. Again the goal of this step is to train the backbone module to extract the visual representation of cell images. 
%Note that, to avoid the catastrophic forgetting during this step, \TheName{} leverages $L^2$-SP strategy to regularize the updates to weights of backbone module.

%to make $F$ satisfy the following equation,
%\begin{equation}\label{equ:contastive_satisfies}
 %   s\left(F(I_{1}), F\left(I_{2}\right) \right) \gg s\left(F(I_{1}), F\left(I^{-}\right) \right)
%\end{equation}
%In Equ.~\ref{equ:contra_loss}, $I_{1}, I_{2}$ are positive pair, the $I^{-}\in \Omega^-$ denotes the negative samples of $I$, and the $\cdot$ symbol denotes the dot product and $\tau$ denotes the temperature hyper-parameter.
%And in Equ.~\ref{equ:contastive_satisfies}, $s(\cdot, \cdot)$ is a function that defines the similarity between two images. 

\subsubsection{COCO Adaption Step} Given the updated backbone in the previous step, this step combines the backbone module with the neck and head modules obtained from the previous iterations and reconstructs the full Cascade Mask R-CNN model. Then, it 
leverages \emph{COCO} dataset to preserve the instance segmentation capacity of the neck and head modules, even with the learned representation from the updated backbone.

{\bf Regularization to Catastrophic Forgetting}. Different from \emph{COCO Pre-training} introduced in Section~\ref{sssec:coco pretrain}, 
%this step only back-propagates to update the weights of neck and head modules, while keeping the weights of backbone frozen.
to avoid the catastrophic forgetting, \TheName{} adopts the $L^2$-SP strategy \cite{xuhong2018explicit}.
%
%During the \emph{COCO Adaption Step}, the backbone would ``forget'' the visual domain knowledge of cells learned from the previous \emph{Cell Images MoCo Step}. To solve the problem, \TheName{} leverages a knowledge transfer regularization derived from $L^2$-SP strategy \cite{xuhong2018explicit}. 
%In each iteration of \emph{COCO Adaption Step}, 
Given the weights of pre-trained backbone $\boldsymbol{w}^0_{\mathcal{S}}$ obtained from the previous \emph{Cell Images MoCo step}, \TheName{} regularizes the training procedure of backbone module as follow.
\begin{equation}\label{equ:l2-sp}
   \Omega(\boldsymbol{w}_{\mathcal{S}})=
   \alpha\left\|\boldsymbol{w}_{\mathcal{S}}-\boldsymbol{w}_{\mathcal{S}}^{0}\right\|_{2}^{2}
   +\left(1-\alpha\right)
   \left\|\boldsymbol{w}_{\mathcal{S}}\right\|_{2}^{2}
\end{equation}
where $\boldsymbol{w}_{\mathcal{S}}$ is the learning outcome and $\alpha$ is the hyper-parameter. Thus, above regularization constrains the distance between $\boldsymbol{w}_{\mathcal{S}}$ and the backbone trained by the previous step.

\subsection{Fine-tuning}\label{sssec:finetuning}
Given the pre-trained DNN %using the above two steps
by \emph{COCO Pre-training} and \emph{AMT$^2$ Procedure}
and a few shots of annotated cell images (with bounding boxes, segmentation masks, and cell types for every cell in every image), \TheName{} further trains the whole DNN model, including the backbone, neck, and head modules using the annotated cell images. Specifically, \TheName{} employs the standard instance segmentation loss $\mathcal{L}_{inst}$ defined in Eq.~\ref{equ:instance_loss} for fine-tuning. Please refer to  Sec.~\ref{ssec:few-shot learning setups} for the details on the few-shot learning settings on the two cell image datasets including LIVECell \cite{edlund2021livecell} and BBBC038 \cite{caicedo2019nucleus}.
%\section{Configurations}

%To pre-train DNN, firstly, \TheName{} warmed up DNN model using the labeled \emph{COCO} dataset to learn capability of object detection and instance segmentation descried in~\ref{sssec:coco pretrain}. 

%Then it transferred the capability to domain of cell images via AMT$^2$ described in Sec.~\ref{sssection:amt procedure} using labeled \emph{COCO} \cite{lin2014microsoft} again and unlabeled aggregation of nine cell image datasets, including LIVECell~\cite{edlund2021livecell}, EVICAN~\cite{3.AJBV1S_2019}, DIC-C2DH-HeLa, Fluo-C2DL-MSC, Fluo-N2DH-GOWT1, Fluo-N2DH-SIM+, Fluo-N2DL-HeLa, PhC-C2DH-U373 and PhC-C2DL-PSC. 
%Finally, \TheName{} fine-tuned the DNN descried in~\ref{sssec:finetuning} with a small proportion of annotated cell dataset (e.g. LIVECell~\cite{edlund2021livecell} and BBBC038 \cite{caicedo2019nucleus}).
%Please note that the DNN did not learn any label information of cell image during any pre-training step of \TheName{}.

%In the step of our proposed AMT$^2$ procedure, to observe the design of \TheName{} described in Sec.~\ref{sssection:amt procedure}, \TheName{} splits the \emph{COCO} dataset \cite{lin2014microsoft} into ten equal parts (where each part has a similar number of images as the cell images in the pre-training phase) and repeat the AMT$^2$ for ten rounds.

%% To better present details of our experiments, we reported the number of images/cells in Table~\ref{tab:number-info-of-image/cell} for various experiment settings, including datasets settings, train/valid/test settings, few-shot settings, and single cell-type settings.

\section{Experiment}
We carry out extensive experiments to evaluate the effectiveness of \TheName{}. This section includes the introduction to baseline algorithms for comparisons and the experiment setups to simulate few-shot cell segmentation tasks. We also provide the experiment results and analysis in-depth.

% \usepackage{multirow}
\begin{table*}[]
\centering
\caption{Overview Results of \TheName{} and other Pre-training Algorithm}
\label{tab:overview results}
\renewcommand\arraystretch{0.75}
\renewcommand\tabcolsep{3.0pt}
% \resizebox{0.8\textwidth}{45mm}{
\begin{tabular}{lllrrrrrrrr}
\toprule
\multirow{2}{*}{\textbf{Datasets}} & \multirow{2}{*}{} & \multirow{2}{*}{} & \multicolumn{4}{c}{$\mathbf{AP_{bbox}}$} & \multicolumn{4}{c}{$\mathbf{AP_{segm}}$} \\
\cmidrule(r){4-7}\cmidrule(r){8-11}
 &  &  & \multicolumn{1}{c}{\textbf{Scratch}} & \multicolumn{1}{c}{\textbf{Cells-MoCo}} & \multicolumn{1}{c}{\textbf{COCO}} & \multicolumn{1}{c}{\textbf{\TheName{}}} & \multicolumn{1}{c}{\textbf{Scratch}} & \multicolumn{1}{c}{\textbf{Cells-MoCo}} & \multicolumn{1}{c}{\textbf{COCO}} & \multicolumn{1}{c}{\textbf{\TheName{}}} \\
 \midrule
\multirow{17}{*}{LIVECell \cite{edlund2021livecell}} & \multirow{5}{*}{All} & 5\% & 9.9 & 8.3 & 40.0 & \textbf{41.5} & 11.0 & 9.1 & 41.2 & \textbf{42.5} \\
 &  & 10\% & 17.7 & 18.3 & 43.5 & \textbf{44.0} & 19.0 & 19.3 & 43.9 & \textbf{44.6} \\
 &  & 15\% & 17.8 & 23.7 & 44.9 & \textbf{45.0} & 18.5 & 24.0 & 45.1 & \textbf{45.3} \\
 &  & 20\% & 29.7 & 30.8 & 44.8 & \textbf{45.7} & 30.6 & 32.0 & 45.0 & \textbf{46.0} \\
 \cmidrule(r){3-11}
 &  & 100\% & 38.4 & 30.0 & \textbf{47.7} & \textbf{47.7} & 38.6 & 30.4 & 47.8 & \textbf{47.9} \\
 \cmidrule(r){2-11}
 & \multirow{3}{*}{SH-SY-5Y} & 5\% & 0.8 & 1.1 & 16.8 & \textbf{17.4} & 0.9 & 1.1 & 15.7 & \textbf{16.0} \\
 &  & 10\% & 2.3 & 1.2 & 18.9 & \textbf{19.2} & 2.3 & 1.0 & 17.8 & \textbf{18.5} \\
 \cmidrule(r){3-11}
 &  & 100\% & 17.0 & 20.7 & 23.3 & \textbf{25.3} & 15.2 & 19.1 & 21.8 & \textbf{23.8} \\
 \cmidrule(r){2-11}
 & \multirow{3}{*}{MCF7} & 5\% & 6.9 & 6.7 & 29.6 & \textbf{29.9} & 7.8 & 8.0 & 30.9 & \textbf{31.0} \\
 &  & 10\% & 13.7 & 7.3 & 30.7 & \textbf{31.6} & 14.9 & 7.9 & 31.8 & \textbf{32.7} \\
 \cmidrule(r){3-11}
 &  & 100\% & 28.2 & 27.2 & 35.4 & \textbf{36.0} & 29.5 & 28.4 & 36.8 & \textbf{37.1} \\
 \cmidrule(r){2-11}
 & \multirow{3}{*}{A172} & 5\% & 6.3 & 2.0 & 29.5 & \textbf{30.0} & 8.5 & 2.4 & 31.2 & \textbf{32.3} \\
 &  & 10\% & 16.6 & 7.3 & 29.4 & \textbf{32.4} & 19.4 & 8.1 & 31.4 & \textbf{34.9} \\
 \cmidrule(r){3-11}
 &  & 100\% & 25.1 & 23.5 & \textbf{35.9} & 35.7 & 27.0 & 25.5 & \textbf{37.7} & 37.5 \\
 \cmidrule(r){2-11}
 & \multirow{3}{*}{BT474} & 5\% & 8.6 & 9.1 & 32.3 & \textbf{33.4} & 9.1 & 9.7 & 33.3 & \textbf{34.0} \\
 &  & 10\% & 20.7 & 17.5 & \textbf{35.1} & 34.8 & 21.6 & 18.0 & 35.6 & \textbf{36.0} \\
 \cmidrule(r){3-11}
 &  & 100\% & 38.9 & 34.9 & 39.9 & \textbf{42.2} & 40.0 & 36.0 & 39.9 & \textbf{43.0} \\
 \midrule
  \multirow{5}{*}{BBBC038 \cite{caicedo2019nucleus}} & \multirow{5}{*}{All}  & 5\% & 1.0 & 0.5 & 43.7 & \textbf{53.8} & 0.7 & 0.3 & 37.5 & \textbf{45.8} \\
 &  & 10\% & 4.7 & 4.5 & 41.4 & \textbf{60.7} & 3.3 & 3.1 & 35.1 & \textbf{51.6} \\
 &  & 15\% & 18.0 & 6.5 & 59.5 & \textbf{60.4} & 15.1 & 6.4 & 51.0 & \textbf{51.9}  \\
 &  & 20\% & 15.1 & 13.9 & 60.9 & \textbf{62.1} & 16.0 & 12.8 & 52.3 & \textbf{54.0} \\
 \cmidrule(r){3-11}
 &  & 100\% & 26.1 & 22.2 & 62.0 & \textbf{62.6} & 22.3 & 17.5 & 52.3 & \textbf{53.3}\\
 \bottomrule
\end{tabular}
% }
\end{table*}

\subsection{Baselines Algorithms}

We present several pre-training algorithms or initialization strategies for comparison as follows.
\begin{itemize}
    \item \textbf{Scratch}: The models are all initialized using Kaiming's strategies \cite{he2015delving} and fine-tuned on the target datasets;
    \item \textbf{COCO} pre-trained models: The models are initialized using the officially-released weights pre-trained by the \emph{COCO} \cite{lin2014microsoft} dataset and fine-tuned on the target datasets;
    \item \textbf{Cells-MoCo} pre-trained models: The backbone of model is pre-trained using \emph{MoCo} on cell datasets, and neck and head modules are initialized using Kaiming's random initialization \cite{he2015delving}.
\end{itemize}
All the above methods are used for performance evaluation and comparisons under fair comparison settings accordingly.

\begin{figure*}%[htb]
\centering
\includegraphics[width=\textwidth]{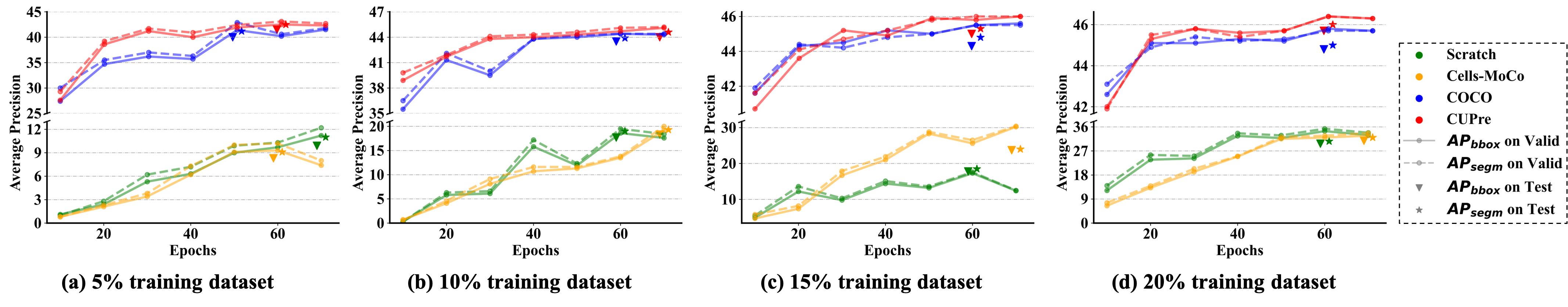}
\caption{$AP_{bbox}$ and $AP_{segm}$ versus the number of training epochs on LIVECell \cite{edlund2021livecell} with 5\%, 10\%, 15\%, and 20\% training data size. \TheName{} clearly outperforms the \emph{COCO} \cite{lin2014microsoft} pre-training algorithm with any training data size setting.}
\label{fig:livecell-few-shot-learning}
\vspace{-5mm}
\end{figure*}

\subsection{Few-shot Learning Setups}\label{ssec:few-shot learning setups}
In this section, we introduce the few-shot learning dataset preparation and evaluation metrics in the experiments. 

{\bf Few-shot LIVECell}. While the LIVECell~\cite{edlund2021livecell} dataset is actually an large annotated cell image dataset (with 5,387 images annotated), we here use part of cell images and annotations from LIVECell to simulate few-shot cell segmentation tasks. Specifically, we randomly pick up 5\%, 10\%, 15\% and 20\% annotated cell images from the training dataset of LIVECell to formulate few-shot cell segmentation tasks for evaluation and analysis.
%
%performed few-shot learning on LIVECell \cite{edlund2021livecell} dataset for \TheName{} with 5\%/10\%/15\%/20\% training data size setting, and made an evaluation on single cell-type experiments of LIVECell \cite{edlund2021livecell} with the same few-shot learning settings. 
%
To better demonstrate the performance of \TheName{}, we  selected four single-cell types for in-depth analysis, including SH-SY-5Y, MCF7, A172 and BT474. Those cells are the four most difficult objects to be detected out of the eight types, as observed in work by \citeauthor{edlund2021livecell}~\cite{edlund2021livecell}.

{\bf External Validation with BBBC038}. To further evaluate the transfer capability of model pre-trained by \TheName{}, we also evaluate the performance of \TheName{} on the BBBC038 \cite{caicedo2019nucleus} --- an additional dataset that never appears in the pre-training phase, to perform an external validation. 
More importantly, compared to LIVECell \cite{edlund2021livecell}, the size of the BBBC038 \cite{caicedo2019nucleus} dataset is relatively small and data distributions are different. Specifically, BBBC038 only contains 670 images and BBBC038 is with only 45 cells per image, versus LIVECell \cite{edlund2021livecell} with 313 cells per image, on average. For the external validation, \TheName{} randomly selected 80\% images of the BBBC038 \cite{caicedo2019nucleus} dataset for fine-tuning and utilized the rest for validation and testing.

%\subsection{Evaluation Metrics}
To evaluate the performance of the proposed \TheName{}, we observe standard practices from the Microsoft \emph{COCO} evaluation protocol \cite{lin2014microsoft} to evaluate cell detection and segmentation quality in the instance segmentation task, and we followed modifications reported in Edlund et al. \cite{edlund2021livecell} to better reflect cell sizes. 
For our evaluation metrics, we report the overall \emph{\underline{A}verage \underline{P}recision of \underline{bbox} detection} ($AP_{bbox}$) and \emph{\underline{A}verage \underline{P}recision of \underline{segm}entation} ($AP_{segm}$) as the overall scores provide a more extensive and rigorous assessment of model performance, where Average Precision (AP) is the precision averaged across all unique recall levels.

%In our experiment, we followed all the above mentioned rules and evaluated \TheName{} on LIVECell \cite{edlund2021livecell} and BBBC038 \cite{caicedo2019nucleus} datasets. We used the training sets of datasets to fine-tune the DNN with \TheName{} and other baseline algorithms, selected the optimal parametric model using validation set, and reported results on validation and testing sets. In addition, for experiments using 100\% of the training dataset, we fine-tuned DNN with 35 epochs and evaluated it on the validation set per 5 epochs for model selection, and for experiments with few-shot learning, we fine-tuned it with 70 epochs and evaluated it per 10 epochs.

%\subsection{Experimental Results}

\subsection{Overall Performance Comparisons}\label{sssec:overall_results}

Table.~\ref{tab:overview results} presents performance of test set using \TheName{} and other pre-training algorithms on LIVECell \cite{edlund2021livecell} and BBBC038 \cite{caicedo2019nucleus} datasets. 
In addition, Table.~\ref{tab:overview results} reports results of experiments with few-shot settings on the two datasets, and provides performances of single cell-type experiments on LIVECell \cite{edlund2021livecell} dataset. 
For details, the Table.~\ref{tab:overview results} shows that \TheName{} achieves the best performance in most results, especially with the few-shot learning setting, \TheName{} gains significant improvements. Only on the experiment of BT474 cells with few-shot setting of 10\%, \TheName{} is slightly lower than \emph{COCO} pre-training algorithms and gets a solid performance on $AP_{bbox}$ metric (34.8\% VS 35.1\%), but is higher than it on $AP_{segm}$ metric (36.0\% VS 35.6\%).
Furthermore, Table.~\ref{tab:overview results} shows that the performance of Scratch and Cells-MoCo is much worse than \emph{COCO} pre-trining algorithms and \TheName{}, since part or the whole components of DNN do not learn the corresponding knowledge using Scratch and Cells-MoCo pre-training algorithms. 
For example, for Cells-MoCo, it provides the knowledge to generate cell-relevant representations for the backbone network of DNN based on the contrastive learning loss, but the head and neck network do not learn the knowledge to exploit these representations and lead to a poor performance.

For performance of \TheName{} on BBBC038 \cite{caicedo2019nucleus}, Table~\ref{tab:overview results} shows that \TheName{} outperforms \emph{COCO} \cite{lin2014microsoft} with a marginal improvement when using the 100\% training dataset (\textcolor{red}{0.6\% $\uparrow$} on $AP_{bbox}$); however, under few-shot settings, \TheName{} achieves a significant improvement. For example, compared to \emph{COCO} \cite{lin2014microsoft}, with a 5\% training dataset, \TheName{} achieves an $AP_{bbox}$ of 53.8\% (\textcolor{red}{10.1\% $\uparrow$}) and an $AP_{segm}$ of 45.8\% (\textcolor{red}{8.3\% $\uparrow$}). 
These results, on the one hand, indicate that \TheName{} provides knowledge for transferring to any new cell dataset unconditionally; on the other hand, they demonstrate that compared to \emph{COCO} \cite{lin2014microsoft}, \TheName{} could be better applied to datasets with a smaller number of training samples and fewer instances. Furthermore, similar to the LIVECell dataset \cite{edlund2021livecell}, Scratch and Cells-MoCo achieve poor performance, due to the lack of complete knowledge of the relevant data.

\subsection{Case Studies and Ablation Analysis}
Here, we interpret the performance comparison results through two case studies and analyze the effectiveness of every component in \TheName{}, all based on \emph{LIVECell} Dataset.

\subsubsection{Performance on All Cell Types in LIVECell Dataset} Fig.~\ref{fig:livecell-few-shot-learning} reports performances of $AP_{bbox}$ and $AP_{segm}$ on LIVECell \cite{edlund2021livecell} dataset with 5\%/10\%15\%/20\% training data size using various pre-training algorithms. 
From Fig.~\ref{fig:livecell-few-shot-learning} we can clearly see that \TheName{} outperforms \emph{COCO} \cite{lin2014microsoft} with any data size setting. %, and it is a reliable result due to the performance of testing set. 
In addition, we find that in the learning process with extremely limited labeled data (e.g. 5\% and 10\% data size), \emph{COCO} \cite{lin2014microsoft} presents significant instability, but \TheName{} does not, since \emph{COCO} \cite{lin2014microsoft} was built using a large number of natural images, which are dissimilar to cell images and lead to a bad knowledge transfer.
However, \TheName{} benefits from the self-supervised learning paradigm to learn cell-relevant representations from a large number of cell images, and it thus achieves a stable learning process and obtains higher performance.
Furthermore, these figures suggest Cells-MoCo performed similarly as Scratch.%, which means that it is necessary to provide knowledge for all components of the model.

\begin{figure}%[htb]
\centering
\includegraphics[width=0.5\textwidth]{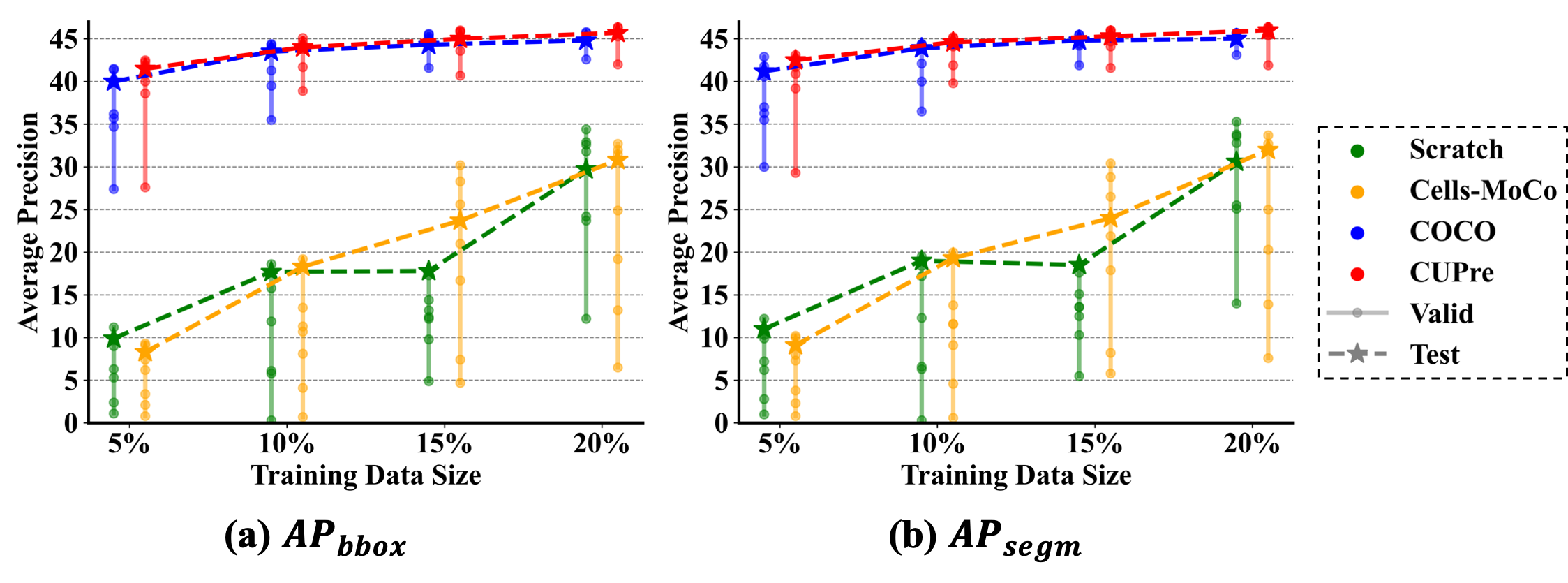}
\caption{$AP_{bbox}$ and $AP_{segm}$ versus the training sample size as described in par. \emph{few-shot learning setting} on LIVECell \cite{edlund2021livecell} via various pre-training methods. \TheName{} outperforms \emph{COCO} \cite{lin2014microsoft} even with fewer training data.} 
\label{fig:livecell-few-shot-learning-overview}
\vspace{-5mm}
\end{figure}

Fig.~\ref{fig:livecell-few-shot-learning-overview} further summarizes the performance of four pre-training algorithms on the validation and test sets of LIVECell \cite{edlund2021livecell} with various training data size settings. 
From Fig.~\ref{fig:livecell-few-shot-learning-overview} we can see that as more labeled datasets are used for learning, 
DNN achieves a steady performance improvement, regardless of which pre-training algorithm is used. 
A large labeled dataset is significant for the performance improvement of the DNN model. However, based on  representations learned from a larger number of unlabeled datasets, \TheName{} outperforms \emph{COCO} \cite{lin2014microsoft} using fewer labeled datasets; for example, on performance of the testing set, \emph{COCO} \cite{lin2014microsoft} achieves an $AP_{bbox}$ of 44.8\% and an $AP_{segm}$ of 45.0\% with 20\% data size, but \TheName{} achieves an $AP_{bbox}$ of 45.0\% (\textcolor{red}{0.2\%$\uparrow$}) and an $AP_{segm}$ of 45.3\% (\textcolor{red}{0.3\%$\uparrow$}) with 15\% (\textcolor{red}{5\%$\downarrow$}) data size.

\begin{figure*}%[htb]
\centering
\includegraphics[width=0.95\textwidth]{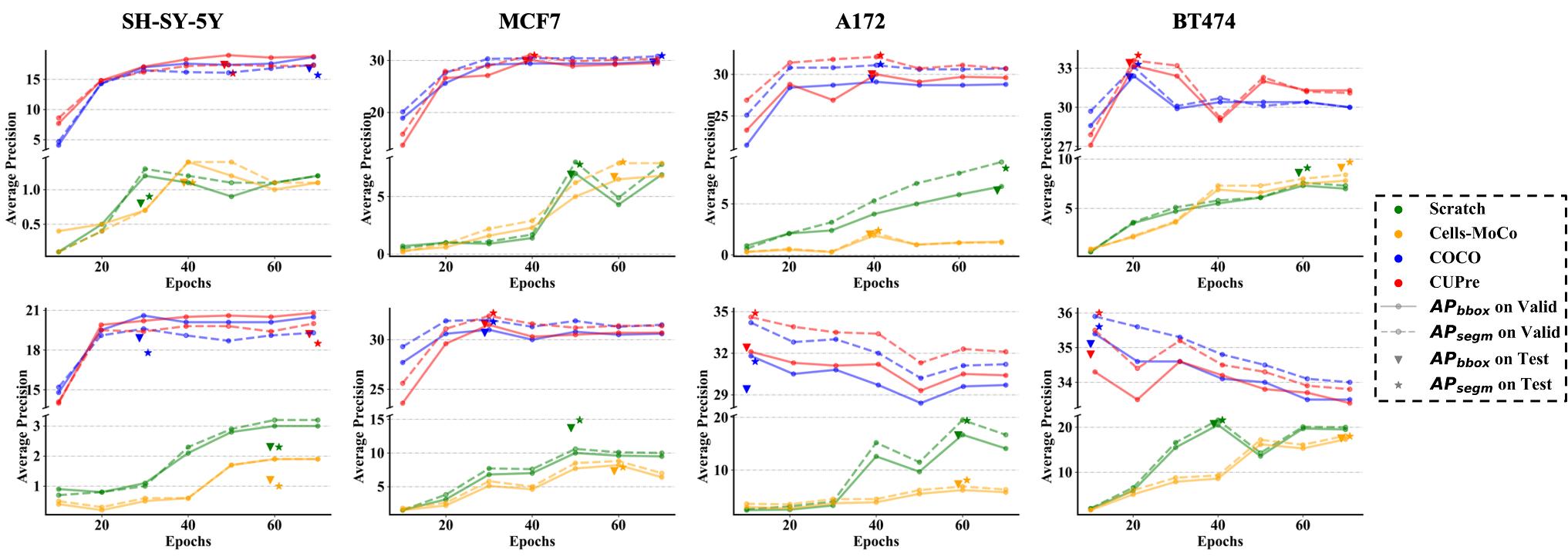}\vspace{-2mm}
\caption{$AP_{bbox}$ and $AP_{segm}$ versus the number of training epochs on four single cell of LIVECell \cite{edlund2021livecell} with 5\% and 10\% training data sizes. }
\label{fig:livecell-single-cell-few-shot-learning}
\vspace{-5mm}
\end{figure*}

\subsubsection{Performance on Every Single Cell-type} Fig.~\ref{fig:livecell-single-cell-few-shot-learning} shows performance of \TheName{} and \emph{COCO} \cite{lin2014microsoft} in four single cell-type experiments. 
It indicates that \TheName{} performed similarly to \emph{COCO} \cite{lin2014microsoft} during training. Specifically, in the experiments with A172 cells, with 5\% of the data, both \TheName{} and \emph{COCO} \cite{lin2014microsoft} achieved optimal performance in the first 20 epochs of learning, and then performed consistently. With 10\% of the data, they both achieved optimal performance in the first 10 epochs and then performed overfitting.
This phenomenon occurs because \TheName{} utilizes a large number of \emph{COCO} \cite{lin2014microsoft} datasets for learning in the first two pre-training phases (\emph{COCO Pre-training} ~\ref{sssec:coco pretrain} and the \emph{AMT$^{2}$ Procedure} ~\ref{sssection:amt procedure}), especially for the neck and head networks of DNN, the knowledge is learned entirely from the \emph{COCO} \cite{lin2014microsoft}. 
But even so, \TheName{} is still able to learn cell knowledge from unlabeled images using contrastive loss and leverages AMT$^{2}$ Procedure to tune neck and head to match representations extracted by the backbone, leading to a higher performance than \emph{COCO} \cite{lin2014microsoft} in most results showed in Fig.~\ref{fig:livecell-single-cell-few-shot-learning}.

\begin{figure}%[htb]
\centering
\includegraphics[width=0.5\textwidth]{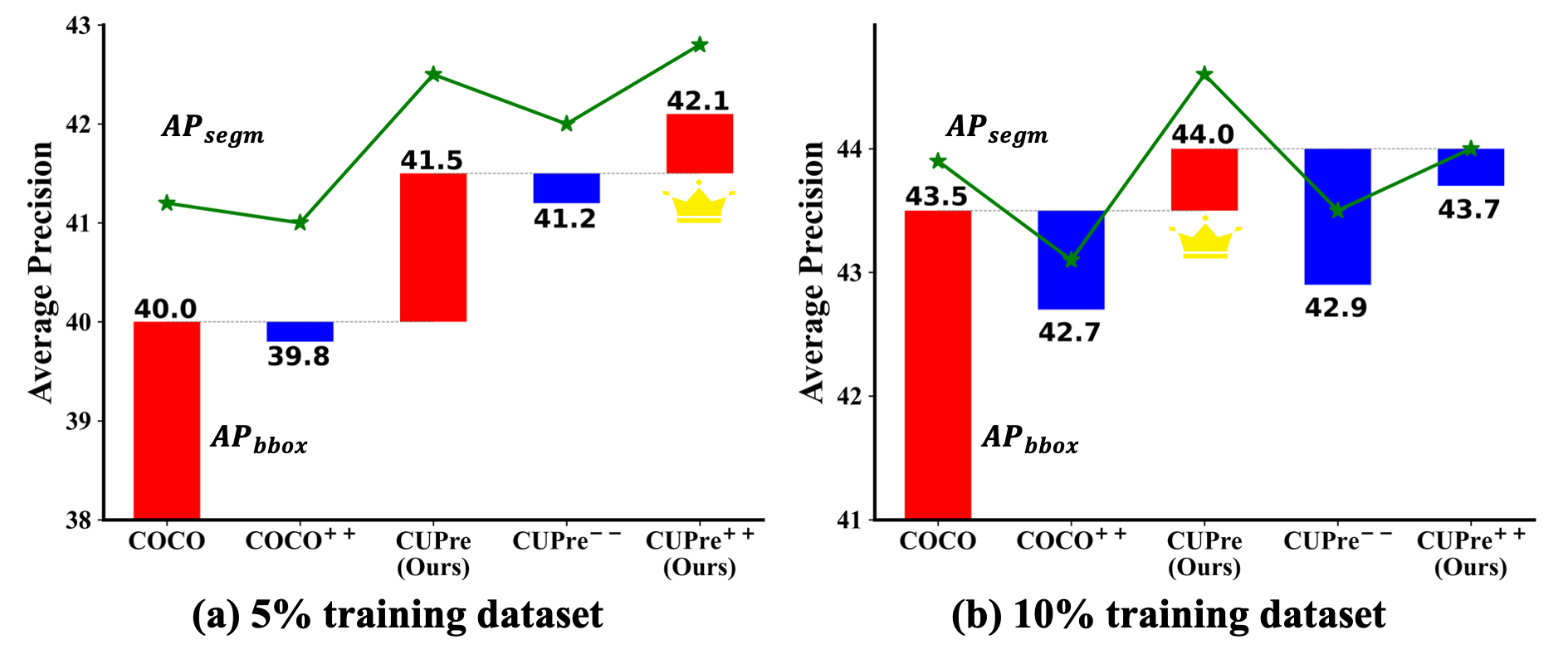}\vspace{-2mm}
\caption{Ablation Studies of \TheName{} performance on the testing sets of LIVECell \cite{edlund2021livecell} dataset, where $AP_{bbox}$ and $AP_{segm}$ values are used as metric. Performance of \TheName{} and all baselines with 5\% and 10\% training dataset size are plotted for ablation studies. On the bar of each subplot, red bars indicate improvements and bars in blue show performance degradation.}
\label{fig:ablation-study}
\vspace{-5mm}
\end{figure}

\begin{figure*}%[htb]
\centering
\includegraphics[width=\textwidth]{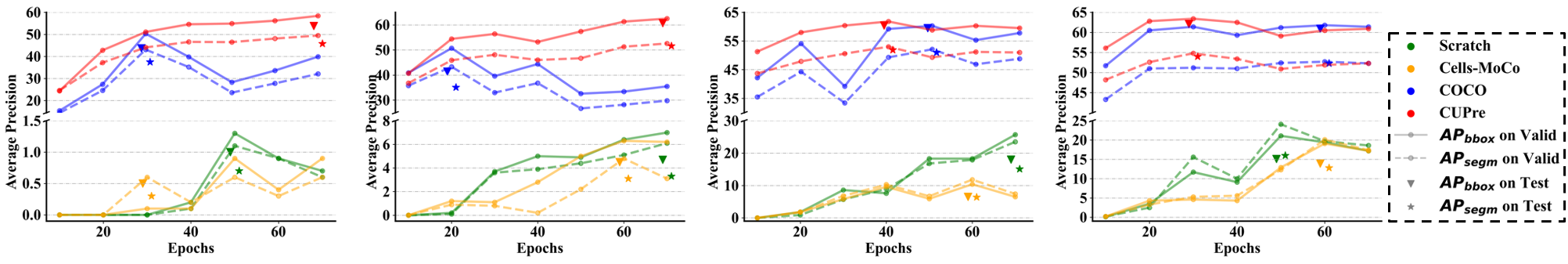}
\caption{$AP_{bbox}$ and $AP_{segm}$ versus the number of training epochs on BBBC038 \cite{caicedo2019nucleus} with 5\%/10\%15\%/20\% training data size. \TheName{} clearly outperforms the \emph{COCO} \cite{lin2014microsoft} pre-training algorithm with any training data size setting.}
\label{fig:BBBC-few-shot-learning}
\vspace{-5mm}
\end{figure*}

\begin{figure}%[htb]
\centering
\includegraphics[width=0.48\textwidth]{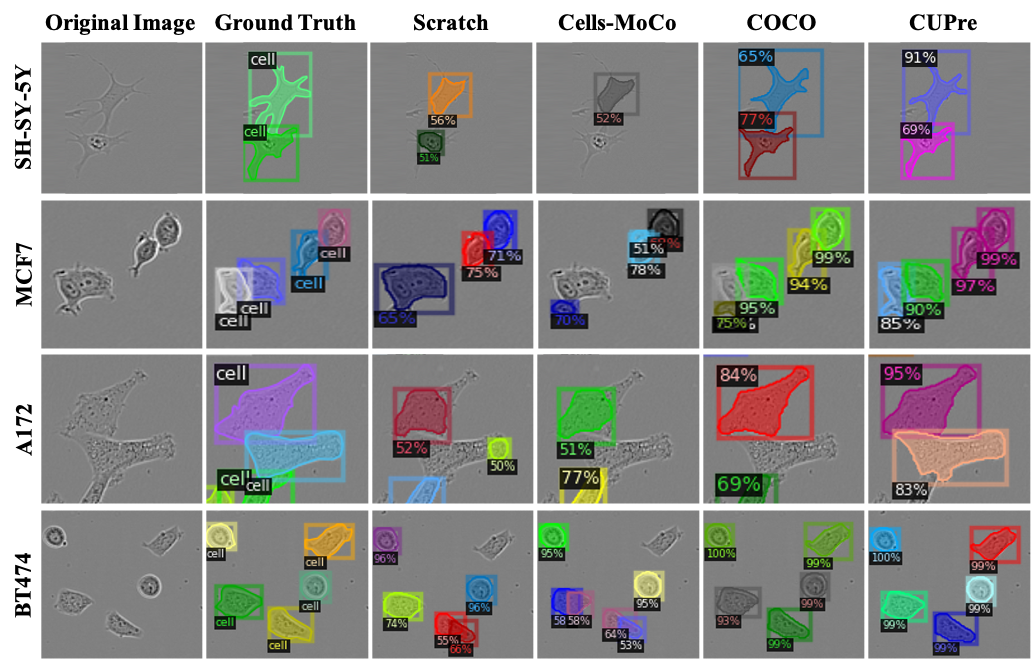}
\caption{Four cases of instance segmentation results using various pre-training algorithms are shown. Each row represents one type of cells in LIVECell. % Gradually decreasing difficulty of cell segmentation from top to bottom.
}
\label{fig:livecell-single-cell-sample-paint}
\vspace{-5mm}
\end{figure}

\begin{figure}%[htb]
\centering
\includegraphics[width=0.48\textwidth]{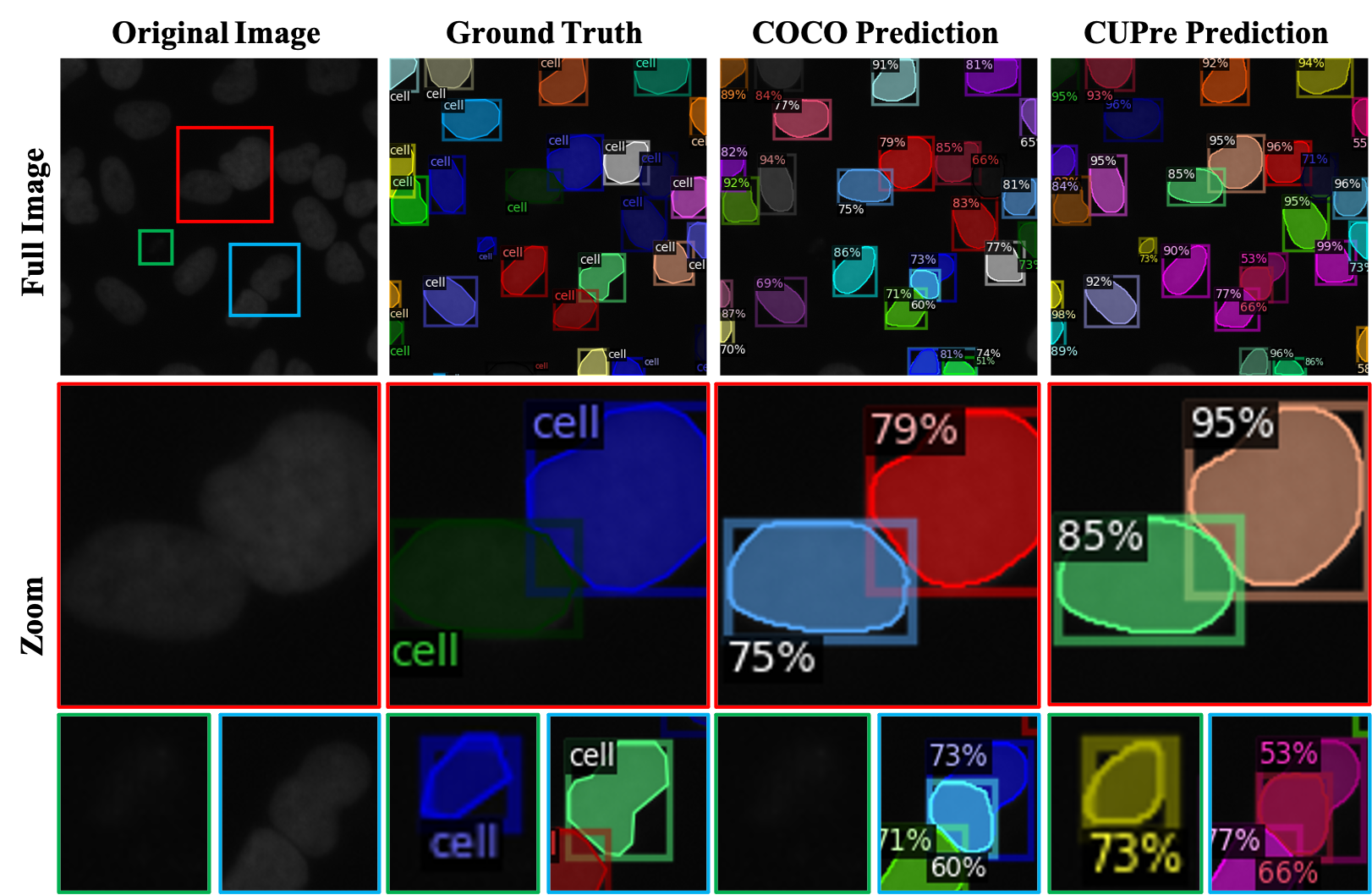}
\caption{Comparison of Prediction using \emph{COCO} \cite{lin2014microsoft} and \TheName{} pre-training algorithm on BBBC038 \cite{caicedo2019nucleus}. %For better visualization, the results of linear normalization (color for gray and viridis) and grayscale histogram for subplot in green area are plotted at bottom.
Please note the subplot with the \textcolor{green}{$\bullet$} box, where the cell is actually existing although difficult to recognize, but could be found by ground truth or some image processing, such as normalization, color mapping.}
\label{fig:BBBC-comparsion-of-coco-cupre}
\vspace{-5mm}
\end{figure}

\begin{figure}%[htb]
\centering
\includegraphics[width=0.92\linewidth]{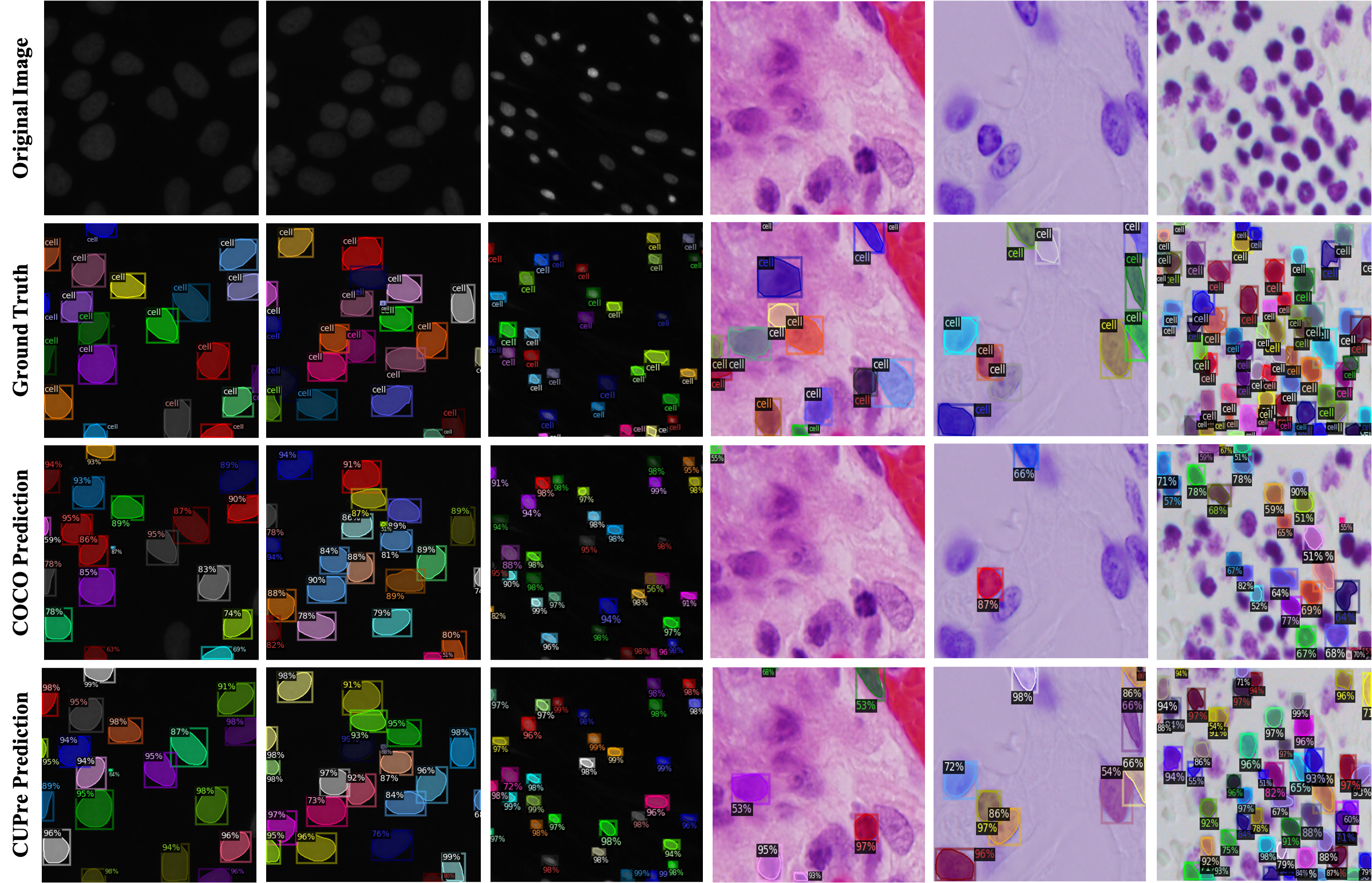}
\caption{Comparison of Prediction using \emph{COCO} \cite{lin2014microsoft} and \TheName{} pre-training algorithm on two types of cells in BBBC038 \cite{caicedo2019nucleus}. The first three columns are gray-scale images and the rest are color images.}
\label{fig:BBBC-comparsion-of-two-types-cells}
\vspace{-5mm}
\end{figure}

\subsubsection{Ablation Studies}To further evaluate the effectiveness of every component incorporated in the pipeline of \TheName{}, we proposed some new baselines, as follows, for comparison.
\begin{itemize}
    \item \textbf{COCO$^{++}$}: The neck and head modules are all initialized using the officially-released weights (pre-trained by the \emph{COCO} \cite{lin2014microsoft} dataset) with a backbone pre-trained on cell images in Table~\ref{tab:dataset} based on \emph{MoCo}, then we fine-tuned the whole model accordingly.
    \item \textbf{\TheName{}$^{--}$}: The model is pre-trained and fine-tuned same as \TheName{}, but without regularization of $L^2$-SP to avoid the catastrophic forgetting during the \emph{AMT$^2$ Procedure}.
    \item \textbf{\TheName{}$^{++}$}: A derivative of \TheName{} that leverages the $L^2$-SP strategy to regularize \emph{Fine-tuning}, so as to avoid the catastrophic forgetting during the \emph{Fine-tuning} procedure.
\end{itemize}

In Fig.~\ref{fig:ablation-study}, we plotted the performance of \emph{COCO} \cite{lin2014microsoft}, \TheName{} and these new proposed pre-training algorithms on the LIVECell \cite{edlund2021livecell} dataset with 5\% and 10\% training dataset size. From it, we clearly know that \textbf{COCO$^{++}$} achieved a worse result than \emph{COCO} \cite{lin2014microsoft} due to mismatched parameters on the backbone, neck, and head. 
\textbf{\TheName{}$^{--}$} achieved a worse result than \TheName{} due to the lack of constraints, leading to overfitting on \emph{COCO} \cite{lin2014microsoft} data without preserving the knowledge of cells learned by MoCo in the previous round.
In addition, \TheName{} achieved the best performance with 10\% training data size, but for only using 5\% train dataset, \textbf{\TheName{}$^{++}$} achieved it. For few-shot learning, \textbf{\TheName{}$^{++}$} prevented overfitting on the training dataset by leveraging L2-SP to preserve knowledge learned during the previous pre-training phase, but when training dataset is enough, e.g. 10\% of LIVECell \cite{edlund2021livecell} data, \TheName{} without any constraints added could better learn information from LIVECell \cite{edlund2021livecell} to achieve better performance.

\subsection{External Validation on BBBC038 Dataset}
To understand the generalizability of \TheName{} on out-of-distribution datasets, we evaluated and compared \TheName{} and other pre-training algorithms on BBBC038 \cite{caicedo2019nucleus} dataset --- a cell dataset that is not used for any pre-training step of \TheName{}.%, which means \TheName{} did not learn any knowledge from this datasets same as \emph{COCO} \cite{lin2014microsoft} pre-training, leading to a fair comparison.
In Fig.~\ref{fig:BBBC-few-shot-learning}, we plot the performance of various pre-training algorithms on the validation set and testing set of BBBC038 \cite{caicedo2019nucleus}, as the same as LIVECell \cite{edlund2021livecell}.
%, record performance of validation set every 10 epochs and evaluate performance of testing set using the optimal checkpoint on the validation set.
We clearly see that \TheName{} outperforms the \emph{COCO} \cite{lin2014microsoft} pre-training algorithm with any training data size setting. 
More importantly, we learned that \emph{COCO} \cite{lin2014microsoft} is not stable in performance on the validation set, especially during learning with few sample images. At 5\% and 10\% data size settings, \emph{COCO} \cite{lin2014microsoft} moves quickly into an overfitting state, and at the 15\% data size setting, it has one more significant fluctuation at 30 epoch. It was not until under the 20\% data size setting that \emph{COCO} \cite{lin2014microsoft} learned steadily. However, in contrast, \TheName{} has a stable performance at any data size setting.
In addition, Scratch and Cells-MoCo again achieve worse results.% as they did on the LIVECell \cite{edlund2021livecell} dataset plotted in Fig.~\ref{fig:livecell-few-shot-learning}.

\subsection{Visualization and Qualitative Results}

In Fig.~\ref{fig:livecell-single-cell-sample-paint} we plotted the instance segmentation result on LIVECell \cite{edlund2021livecell} dataset using \TheName{} and other baseline pre-training algorithms with 5\% training data size. We randomly visualize one sample image for every cell type (totalling 4) we have studied here, and cropped image to demonstrate the local result for the best visualization.

For case of SH-SY-5Y, it is difficult to segment it because of the irregular shape so that DNN provides a viable solution using \emph{COCO} \cite{lin2014microsoft} and \TheName{} but obtains a low confidence score. Using the Scratch and the Cells-MoCo algorithm, the DNN has difficulty capturing the full structure of irregular cells due to the lack of knowledge, and it only performs well in areas with high contrast. For example, the DNN only segments the nucleus in \textcolor[HTML]{183111}{$\bullet$} color area using Scratch initialization.
For case of MCF7, there are two cells that are so close together that we can only identify it due to two nuclei. From segmentation results we see that Scratch regarded it as one cell, Cells-MoCo regarded it as non-cell, \emph{COCO} \cite{lin2014microsoft} regarded it as three cells (Please note the \textcolor[HTML]{75F24F}{$\bullet$}, \textcolor[HTML]{A9A7A9}{$\bullet$} and \textcolor[HTML]{59591C}{$\bullet$} color areas of this subplot.), and only \TheName{} successfully identified it.
For the case of A172, an obscured cell with \textcolor[HTML]{F2AE88}{$\bullet$} color is only segmented by \TheName{}.
For case of BT474, as the easiest of the four cell types to identify, most of the pre-training algorithms achieved a good result, but \TheName{} obtained the most accurate segmentation results and achieved the highest confidence scores.

Fig.~\ref{fig:BBBC-comparsion-of-coco-cupre} shows a visual comparison of the prediction results using \emph{COCO} \cite{lin2014microsoft} and \TheName{} pre-training algorithms. It shows in the red area that when both \emph{COCO} \cite{lin2014microsoft} and \TheName{} algorithms are capable of detecting cells, \TheName{} provides a higher confidence score, and these cells are true positives proved from ground truth. In the green area, the figure suggests that \TheName{} shows stronger performance in the detection of small cells compared to \emph{COCO} \cite{lin2014microsoft}. In addition, in the blue region of results, \TheName{} makes the same mistake as \emph{COCO} \cite{lin2014microsoft}, i.e., mistaking a single cell as two cells.

Fig.~\ref{fig:BBBC-comparsion-of-two-types-cells} shows the prediction results of two different types of cells in the BBBC038 \cite{caicedo2019nucleus} dataset using \emph{COCO} \cite{lin2014microsoft} and \TheName{} pre-training algorithms for instance segmentation. 
In addition to similar findings to Fig.~\ref{fig:BBBC-comparsion-of-coco-cupre}, the figure indicates that a significant experimental finding is that \TheName{} and \emph{COCO} \cite{lin2014microsoft} detected nearly equal numbers of cells in the gray-scale images, although \TheName{} provided a higher confidence score and these cells were true positives. But for color images, \TheName{} detected more cells than \emph{COCO} \cite{lin2014microsoft} and still preserved a high confidence score. The experimental results demonstrate that \TheName{} is able to identify cells in more difficult scenarios compared to \emph{COCO} \cite{lin2014microsoft}.

\section{Limitations and Discussions}
%Though we have made progress in pre-training deep neural networks for few-shot cell segmentation, 
This work still suffers from several major limitations.
First of all, our work focused on pre-training only, while assuming the pre-trained neural networks are further fine-tuned through a standard training procedure of Mask R-CNN~\cite{he2017mask}. In this way, we were not intending to provide an end-to-end optimized solution for cell segmentation, while the overall performance also relies on the way to fine-tune the model pre-trained by \TheName{}. Through extensive experiments, on the top of the Mask R-CNN, we have demonstrated the performance advantages of \TheName{}, compared to the other pre-training strategies under fair comparisons. On the other hand, when advanced fine-tuning strategies or detection \& segmentation modules are incorporated for cell segmentation, the overall performance using \TheName{} for pre-training could be further improved. We would explore more instance segmentation modules for either pre-training or end-to-end cell segmentation in the future.
 
Yet another limitation of \TheName{} is the computational complexity. \TheName{} proposed the \emph{AMT$^2$ Procedure} that requires two alternate pre-training steps, taking more training time than only using self-supervised contrastive learning. Since we just freeze the backbone network during \emph{COCO} pre-training, the speed of AMT$^2$ is satisfying. However, comparisons with other pre-training baselines with varying complexities demonstrated the advantages of \TheName{} in performance improvement, while our ablation studies further confirm the worthiness of every step here. More importantly, the pre-training procedure could be prepared in advance of the acquisition of training data in an offline manner, while pre-trained networks could be reused for new (even out-of-distribution) tasks. The external validations have evaluated the excellent performance of \TheName{} in handling unseen cell images. 
%
%Finally, our studies also show that \TheName{} can work significantly better than other methods in few-shot learning settings, where the number of annotated samples is limited. All these methods in experiments perform similarly when sufficient annotated samples are provided.

%The reason is that more annotated data benefit domain knowledge learning, thus directly fine-tuning \emph{COCO} pre-trained network is able to achieve satisfying performance. Nevertheless, \TheName{} aims to reduce the workload of data annotation, therefore, the performance on few-shot learning is more important.

\section{Conclusions}~\label{sec:conclusion}
In this work, we have presented \TheName{} -- a cross-domain, unsupervised pre-training algorithm that pre-trains deep neural networks using both natural images and cell images.
In particular, \TheName{} learns representations of the visual domain of cells using a contrastive learning paradigm from unlabeled cell images, and it learns representation for the instance segmentation task in a supervised learning manner from the labeled \emph{COCO} \cite{lin2014microsoft} dataset by using the \emph{AMT$^2$ Procedure}, narrowing the gap between two different domains. %Then \TheName{} further proposes AMT$^{2}$ procedure to tackle gap of two different domains. and 
Through pre-training models with the capacity of instance segmentation and good representations of cell images, \TheName{} can adapt to few-shot cell instance segmentation at a low annotation cost.
%Finally, we fine-tuned the network on the cell segmentation task under few-shot learning settings. 
To demonstrate the performance of \TheName{}, we collected ten cell image datasets and one natural image dataset (\emph{COCO}), pre-trained backbone neural networks using these data and \TheName{}, then we fine-tuned the pre-trained backbone on two datasets.
Experiment results on two cell image datasets show \TheName{} outperforms other pre-training methods, including COCO-based and MoCo-based~\ref{sssec:overall_results} solutions. 
More importantly, in our few-shot learning experiment of LIVECell \cite{edlund2021livecell}, \TheName{} outperforms \emph{COCO} \cite{lin2014microsoft} algorithm with fewer datasets (45.0\% $AP_{bbox}$ (\textcolor{red}{0.2\% $\uparrow$}), 45.3\% $AP_{segm}$ (\textcolor{red}{0.3\% $\uparrow$}) and 15\% data size (\textcolor{red}{5\% $\downarrow$})).
Furthermore, we evaluate \TheName{} with one external experiment on BBBC038~\cite{caicedo2019nucleus}, where the datasets of external tasks were never seen in pre-training. 
The excellent performance on the task further confirms the robustness and generalizability of \TheName{} in a variety of cell analytic tasks. 

Note that we design and evaluate \TheName{} for few-shot cell segmentation using the LIVECell~\cite{edlund2021livecell}, which actually is large with more than 3,253 well-annotated cell images in the training set and not for any few-shot tasks. In our work, we just adopt LIVECell to simulate the few-shot settings---a large collection of unlabeled LIVECell images (as well as other unlabeled cell images) used for pre-training, and an extremely small proportion of annotated images used for fine-tuning. However, we believe the proposed method could be also transferred to handle other few-shot cell segmentation tasks from either model-reuse and methodology aspects. The external validation shows that, even when the distributions of two datasets are quite different, the \TheName{} pre-trained model (based on LIVECell) could be still fine-tuned well on BBBC038~\cite{caicedo2019nucleus}. 
%Furthermore, it is reasonable to expect that the proposed \TheName{} pipeline could be easily transferred to unseen few-shot cell segmentation tasks, even when incorporating other datasets for pre-training.
%
%
%the proposed \TheName{} is flexible, and is able to provide better initialization of a variety of backbone, neck and head networks, transferring the capability of instance segmentation learned from \emph{COCO} dataset to few-shot cell segmentation tasks. Hence it can fast adapt to downstream tasks and other cell image datasets, in particular using a few annotated images. 
%
%In the future, we plan to collect an even larger collection of cell images and incorporate both labeled and unlabeled images for pre-training. We would investigate novel \emph{pre-training and fine-tuning} paradigms for cell segmentation under both few-shot or even zero-shot learning settings. 

%it is believed that the \emph{pre-training and fine-tuning} paradigm should be further investigated in the field of cell images and cell segmentation based on various types of microscopy, referring to the domain-shit problem should be considered as a future direction.

\begin{refcontext}[sorting = none]
\printbibliography
\end{refcontext}

\end{document}